\documentclass[a4paper, oneside, twocolumn, notitlepage, 10pt]{extarticle_ecoc}
\usepackage{ecoc}
\usepackage{tabularx,booktabs}
\usepackage{subfigure}
\begin{document}
\selectlanguage{english}    


\title{Experimental Validation of Spectral-Spatial Power Evolution Design Using Raman Amplifiers}%


\author{
   Mehran Soltani\textsuperscript{(1)}, Francesco Da Ros\textsuperscript{(1)},
    Andrea Carena\textsuperscript{(2)}, Darko Zibar\textsuperscript{(1)}
}

\maketitle                  


\begin{strip}
 \begin{author_descr}

    \textsuperscript{(1)}Department of Electrical and Photonics Engineering, Technical University of Denmark, \color{blue}\underline{\href{mailto:msolt@dtu.dk}{msolt@dtu.dk}} \color{black}
    
   \textsuperscript{(2)}Department of Electronics and Telecommunications, Politecnico di Torino, Italy

 \end{author_descr}
\end{strip}

\setstretch{1.1}
\renewcommand\footnotemark{}
\renewcommand\footnoterule{}


\begin{strip}
  \begin{ecoc_abstract}
    We experimentally validate a machine learning-enabled Raman amplification framework, capable of jointly shaping the signal power evolution in two domains: frequency and fiber distance. The proposed experiment addresses the amplification in the whole C-band, by optimizing four first-order counter-propagating Raman pumps. \textcopyright2022 The Author(s)

  \end{ecoc_abstract}
\end{strip}


\section{Introduction}

Distributed Raman amplification is extensively studied as it offers several advantages over the lumped amplifiers such as providing a lower noise figure, broad gain range and higher flexibility in design\cite{Agrawal, 7163281}. One approach in designing Raman amplifiers is to optimize the pump parameters to obtain a desired signal power evolution shape, jointly in spectral and spatial (fiber distance) domains. Controlling the signal power evolution in the frequency and distance, as a two-dimensional (2D) space, is a beneficial way to approach some of the long-term goals in optical communication systems such as signal-to-noise (SNR) enhancement and nonlinearity mitigation \cite{PhysRevLett.101.123903, tan2018distributed}. For instance, a flat 2D profile in frequency and distance, resembling a lossless link, minimizes the accumulated amplified spontaneous emission (ASE) noise \cite{Ania-Castanon:04, 1601058, PhysRevLett.101.123903}. This flat 2D profile is also a requirement for the transmission based on Nonlinear Fourier Transform (NFT) \cite{Mollenauer:88, le2015nonlinear}. A 2D symmetric power profile with respect to the middle point in distance is another practical example utilized to mitigate the nonlinear impairments using optical phase conjugation (OPC) systems \cite{Phillips:14, tan2018distributed}.

Power profiles in a 2D space are mostly addressed by heuristic optimization of the Raman pump parameters, without providing a general design framework \cite{1561354, PhysRevLett.101.123903, Rosa1515, Bednyakova:13}. In \cite{Soltani:21, 9721641}, we presented and numerically validated a machine learning framework to optimize Raman pump powers values, targeting power evolution design jointly in frequency and fiber distance. The proposed approach consists of a convolution neural network (CNN) \cite{Soltani:21}, to predict the pump powers values for a desired 2D power profile, followed by differential evolution (DE) \cite{9721641}, as a fine-tuning technique. 

In this paper, we experimentally validate the CNN model and the CNN-assisted DE framework presented in \cite{Soltani:21, 9721641}. In the proposed amplifier setup, signal power evolution is designed jointly in the whole C-band and along the fiber distance using four counter-propagating Raman pumps. The CNN is trained and evaluated with the 2D profiles generated by probing the setup with different pump power values. The predicted pump power values by the CNN for the test profiles result in low maximum absolute error (MAE) values on average, while showing high MAE($>$1 dB) for roughly 2\% of them. To improve the CNN accuracy on the profiles with high MAE values, the DE technique is utilized to fine-tune the pump power values, in real-time on the setup.
\section{Experimental setup}

\begin{figure*}[t]
   \centering
    \includegraphics[width=0.9\linewidth]{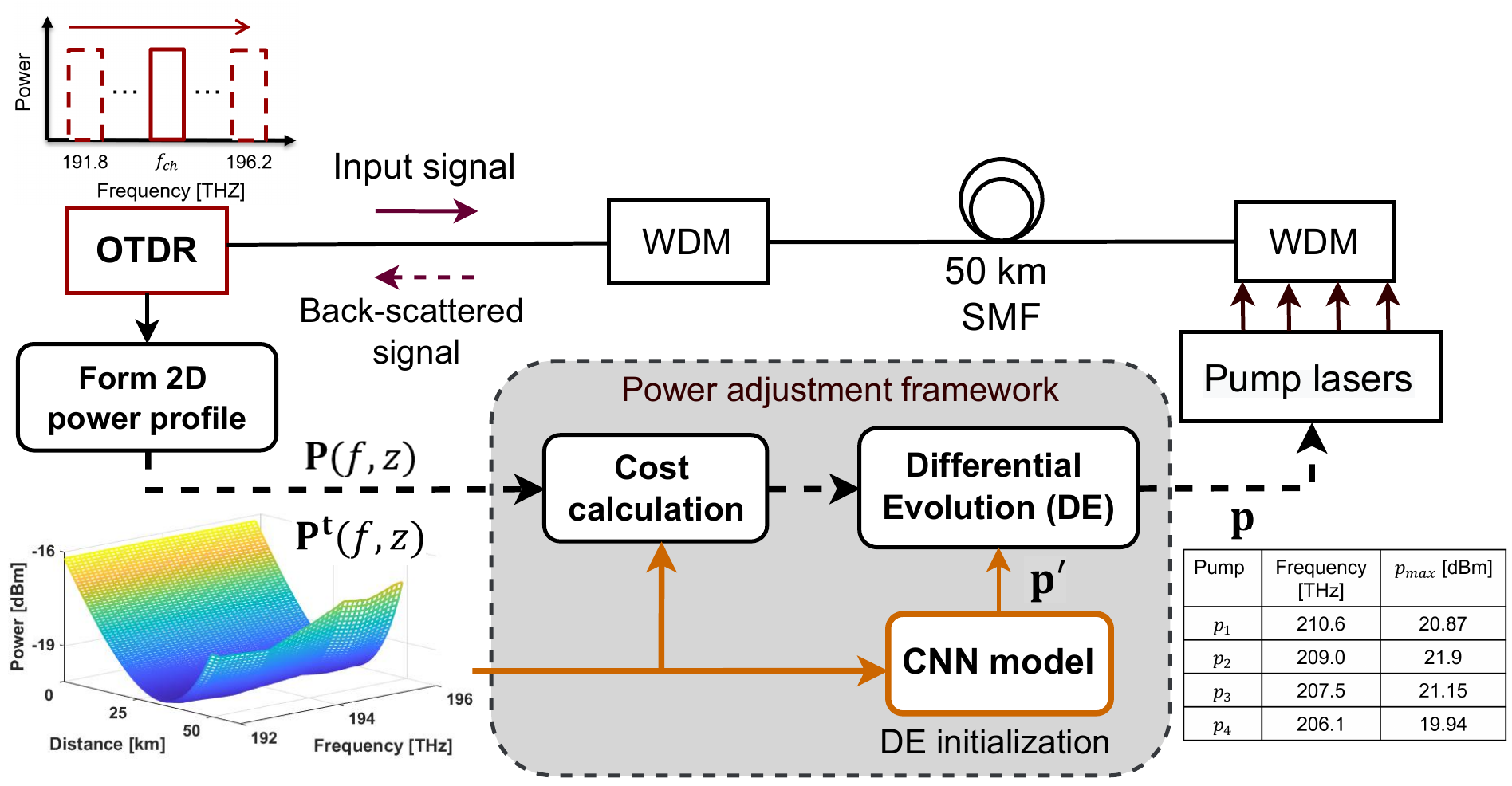}
    \caption{The experimental setup used to adjust the pump powers values for designing a 2D target power profile $\textbf{P}^\textbf{t}(f,z)$.}
    \label{fig:figure2}
\end{figure*}

The experimental setup for the proposed power adjustment framework is shown in Fig.\ref{fig:figure2}. The objective of the framework is to control a set of Raman pump powers values  $\textbf{p} = [p_1, p_2, p_3, p_4]$ to achieve a 2D target profile  $\textbf{P}^\textbf{t}(f,z)$, defined in both spectral (f) and spatial (z) domains. A standard single-mode (SMF) fiber with 50 km length is investigated and the Raman pump module consists of four counter-propagating pump lasers. Pump wavelengths are fixed (shown with their maximum available power value $p_{max}$ in Fig. \ref{fig:figure2}), and able to amplify the whole C-band.



The investigated signal bandwidth covers the C-band between 191.8 THz and 196.2 THz, divided into 44 channels with 100 GHz spacing. To measure the signal power evolution along the fiber distance, a frequency-tunable optical time-domain reflectometer (OTDR) is used. The OTDR is connected to the fiber span with a wavelength division multiplexer (WDM), used to isolate the OTDR from the pump frequencies in the range between 203.9 THz and 211.1 THz. A WDM coupler is placed also at the end of the fiber link to combine the signal and the pumps. 

The OTDR introduces a low power signal (-16 dBm) into the channels, and measures the back-scattered signal in each one of them, sequentially. To reduce the trace noise, the OTDR distance resolution is set to 8.2 m and the signal pulse width is 3 $\mu s$. Once the OTDR measures the signal power evolution for all channels, a Savitzky-Golay smoothing filter \cite{ac60214a047} with the window size $w=19$ (equivalent to 19$\times$8.2=155.8 m) and the polynomial order $n=2$ is applied in distance to reduce the signal fluctuations. The smoothed traces are then down-sampled to achieve 500 m distance resolution, according to the distance resolution values reported in \cite{Soltani:21, 9721641}, and a 2D power profile $\textbf{P}(f,z)$ is formed. $\textbf{P}(f,z)$ is used further as the input to the power adjustment framework to update the pump power values accordingly.


\section{Power adjustment framework}

The power adjustment framework consists of a CNN model followed by DE, known as a gradient-free optimization technique. The CNN learns the mapping between 2D power profiles and their corresponding pump powers, and its model is identical to the one presented in \cite{9721641}. To train the CNN, a data-set is built using the setup by applying randomly selected set of pump powers values and measuring their corresponding 2D profiles. The CNN is trained offline, providing low MAE on average. To further improve the CNN accuracy on a test profile with MAE higher than a threshold (such as 1 dB), DE is used to fine-tune the pump powers values. This fine-tuning is performed in real-time employing the amplifier setup. For a 2D target $\textbf{P}^\textbf{t}(f,z)$, the set of pump powers $\textbf{p}'$ predicted by the CNN is used to initialize the DE population. For each generated individual in the population, the DE process is performed and a new set of pump powers $\textbf{p}$ is applied into the experimental setup (details regarding the DE process can be found in \cite{9721641}). The MAE between the resulting 2D profile $\textbf{P}(f,z)$ and the target 2D profile $\textbf{P}^\textbf{t}(f,z)$ is calculated by considering the maximum error over the both dimensions. MAE is chosen as the cost function for the DE to update the pump power values. The DE process continues until a convergence criteria such as maximum number of iterations or a minimum MAE level is achieved. 

\section{Results}

\begin{figure}[t]
   \centering
        \includegraphics[width=\linewidth]{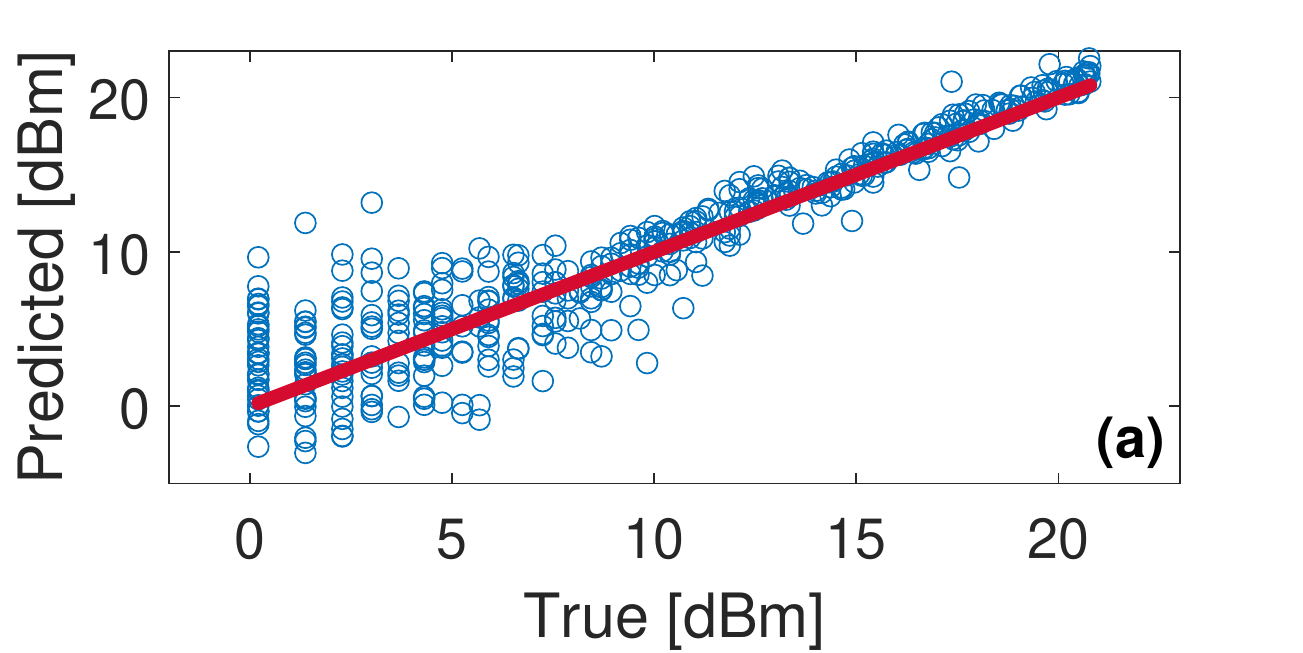}
    \caption{True vs. predicted values for $p_1$.}
    \label{true predicted}
\end{figure}

The first evaluation step is to train and validate the CNN model. We have collected a data-set using the setup, including 4900 samples with randomly selected pump powers and their corresponding 2D power profiles. The data-set is divided into train, test and validation sets with 4100, 500 and 300 samples, respectively. In the training set, different subsets with sizes from 1500 to 4100 are investigated to train the CNN and evaluate the resulting validation set accuracy. According to these analyses, 3700 is selected as the final training size. Once the CNN is trained, the $R^2$ score is calculated for the test data to evaluate the correlation between the true and predicted pump power values. The $R^2$ score takes the values between 0 and 1 where the highest value indicates a perfect prediction. The $R^2$ score attained for each pump is reported in Table  \ref{r values}, confirming the good performance of the CNN model in mapping the 2D profiles to their corresponding pump powers values. The CNN performance in terms of true and predicted pump power values is shown in Fig.\ref{true predicted} for the pump with the lowest accuracy ($p_1$). The error is high for low pump power values due to its low impact on the signal profile. As the pump power increases, it becomes more influential and the prediction accuracy increases, consequently.

\begin{table}[h]
\caption{$R^2$ test scores for the CNN model prediction.}
\label{r values}
\begin{tabular}{ | m{1.75cm} | m{0.9cm}| m{0.9cm} | m{0.9cm}| m{0.9cm} | } 
     \hline
    Pump  & $p_1$ & $p_2$ & $p_3$ & $p_4$ \\ \hline
    $R^2$ & 0.86 & 0.87 & 0.91 & 0.93\\ [3pt]
     \hline
\end{tabular}
\end{table}

\begin{figure}[t]
   \centering
        \includegraphics[width=\linewidth]{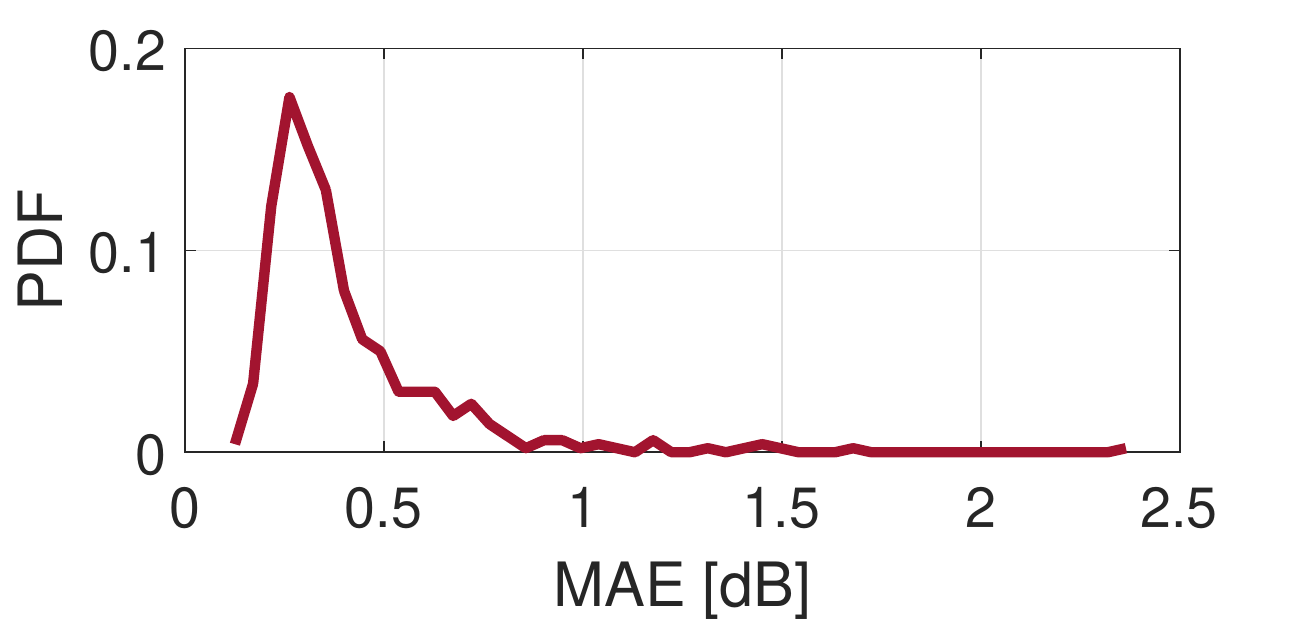}
    \caption{PDF of the MAE for the test data.}
    \label{fig:figure3}
\end{figure}

As an alternative way to evaluate the prediction accuracy, for each target test profile, the predicted pump power values by the CNN are applied to the setup and the MAE between the target 2D profile and the resulting one is calculated. Fig.\ref{fig:figure3} shows the probability density function (PDF) of the MAE for all test profiles, achieving the mean $\mu = 0.37$ dB, and the standard deviation $\sigma = 0.23$ dB. 

\begin{figure}[ht]
   \centering
        \includegraphics[width=\linewidth]{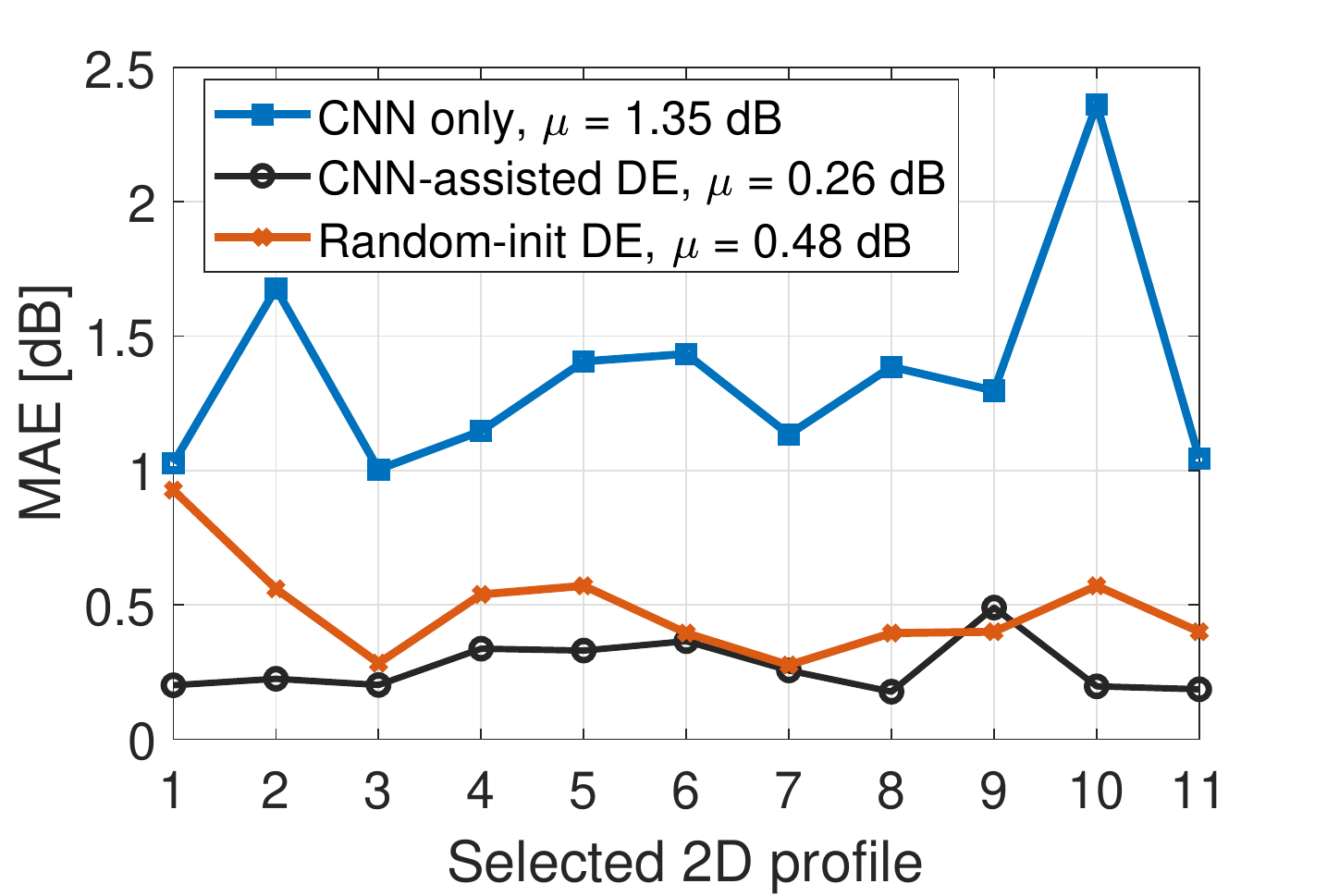}
    \caption{MAE for the CNN, the CNN-assisted DE and the Random-init DE approaches for the selected 2D profiles.}
    \label{fig:figure4}
\end{figure}

\begin{figure}[!]
   \centering
        \includegraphics[width=\linewidth]{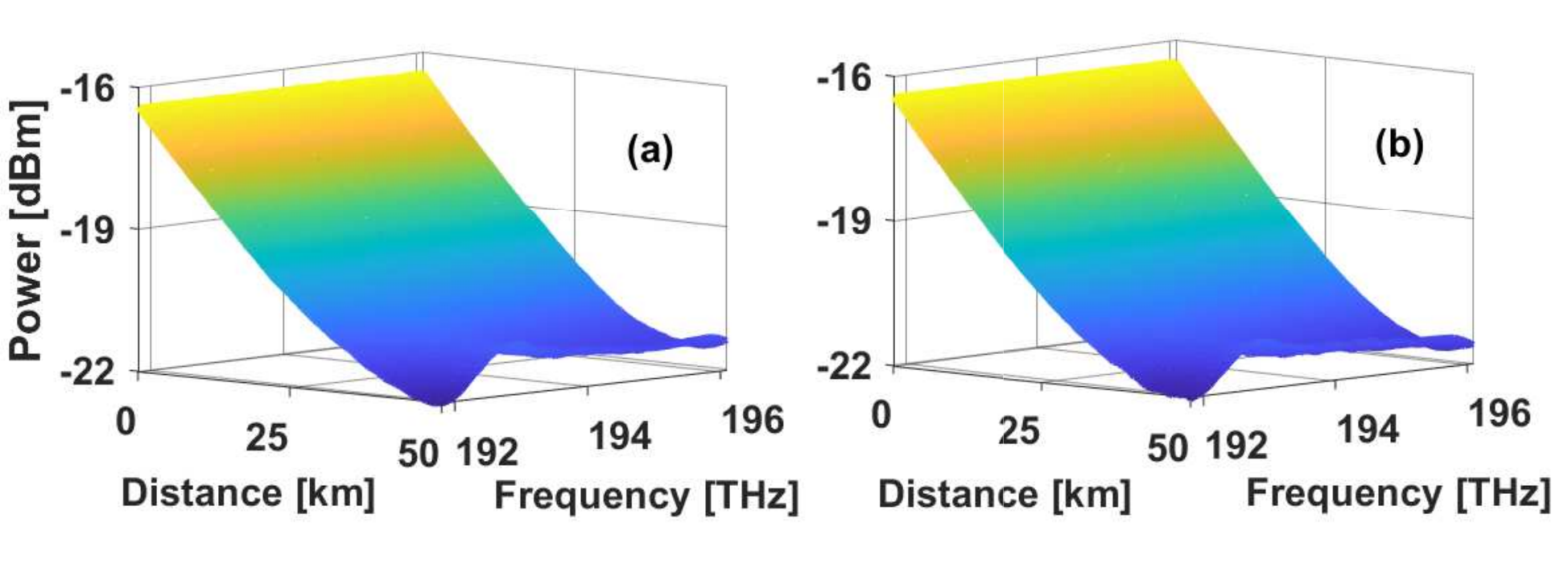}
    \caption{CNN-assisted DE result for the 10th selected 2D profile in Fig.\ref{fig:figure4}. (a) Target 2D profile, (b) Resulting 2D profile. }
    \label{fig:figure5}
\end{figure}

The CNN obtains a low MAE statistically, while for eleven samples, it results in MAE higher than 1 dB. To improve the accuracy for the eleven profiles with high MAE values, we apply the CNN-assisted DE framework as depicted in Fig. \ref{fig:figure2}. The CNN prediction for each selected 2D profile is used to initialize the DE with 100 maximum number of iterations. The DE parameters are set as reported in \cite{9721641}. To show the CNN impact on DE performance, another set of experiments for these eleven profiles is performed, where the DE population is initialized randomly (Random-init DE), without involving the CNN. For each of these selected 2D profiles, indexed from 1 to 11, the MAE is shown in Fig.\ref{fig:figure4} for the CNN only, the CNN-assisted DE and the Random-init DE scenarios. The resulting error using CNN-assisted DE for all eleven samples is less than 0.5 dB in all cases (less than 0.4 dB in 10 out of 11 2D profiles), considerably better than the CNN only and the Random-init DE results. Based on \cite{9721641}, improvement of CNN-assisted DE over the regular Random-init DE would be more significant in case the search space has high number of dimensions, i.e. number of the pumps. To have an intuition on the results achieved by the CNN-assisted DE framework, a sample 2D profile (10th selected 2D profile in Fig.\ref{fig:figure4}) and the actual resulting one, after applying the CNN-assisted DE, is shown in Fig.\ref{fig:figure5}.

\section{Conclusions}
The CNN-assisted DE framework is experimentally validated for designing 2D power evolution profiles using Raman amplifiers. The CNN model achieves less than 0.4 dB test error on average while it is inaccurate for eleven 2D target profiles in the test data-set. Addressing these profiles, DE is applied to fine-tune the pump powers values, showing more than 1 dB average improvement over the CNN. The results assert that the proposed framework can be effectively used to design 2D profiles, by real-time optimization of the amplifier parameters in a setup under test.



\section{Acknowledgements}

This work was supported by the European Research Council (ERC-CoG FRECOM grant no. 771878), the Villum Foundation (OPTIC-AI grant no. 29334), and the Italian Ministry for University and Research (PRIN 2017, project FIRST).


\printbibliography

\vspace{-4mm}

\end{document}